\documentclass[letterpaper, 10 pt, conference]{ieeeconf}  

\IEEEoverridecommandlockouts                              

\usepackage{times}
\usepackage{multicol}
\usepackage[bookmarks=true]{hyperref}
\usepackage{amsfonts}
\usepackage{amsmath,bm}
\usepackage{amssymb}
\usepackage{CJK}
\usepackage{indentfirst}
\usepackage{cases}
\usepackage{algorithm}
\usepackage[algo2e]{algorithm2e}
\usepackage{graphicx}
\usepackage{stfloats}
\usepackage{subfigure}
\usepackage{booktabs}
\usepackage{color,soul} 
\definecolor{red}{rgb}{1.00,0.00,0.00}
\definecolor{blue}{rgb}{0.00,0.00,1.00}
\definecolor{green}{rgb}{0.30, 0.50,0.00}

\newcommand{\cblue}[1] {\textcolor{blue}{#1}}

\usepackage{flushend} 
\usepackage{arydshln}
\usepackage{color}
\usepackage{tikz}
\usepackage{cite}

\title{\LARGE \bf Learning Dual-Arm Coordination for Grasping Large Flat Objects}

\author{Yongliang Wang, Hamidreza Kasaei
\thanks{The authors are with the Department of Artificial Intelligence, Bernoulli Institute, Faculty of Science and Engineering, University of Groningen, The Netherlands
        {\tt\small \{yongliang.wang, hamidreza.kasaei\}@rug.nl}}%
}

\begin{document}

\maketitle
\thispagestyle{empty}
\pagestyle{empty}

\begin{abstract}


Grasping large flat objects, such as books or keyboards lying horizontally, presents significant challenges for single-arm robotic systems, often requiring extra actions like pushing objects against walls or moving them to the edge of a surface to facilitate grasping. In contrast, dual-arm manipulation, inspired by human dexterity, offers a more refined solution by directly coordinating both arms to lift and grasp the object without the need for complex repositioning. In this paper, we propose a model-free deep reinforcement learning (DRL) framework to enable dual-arm coordination for grasping large flat objects. We utilize a large scale grasp pose detection model as a backbone to extract high-dimensional features from input images, which are then used as the state representation in a reinforcement learning (RL) model. A CNN-based Proximal Policy Optimization (PPO) algorithm with shared Actor-Critic layers is employed to learn coordinated dual-arm grasp actions. The system is trained and tested in Isaac Gym and deployed to real robots. Experimental results demonstrate that our policy can effectively grasp large flat objects without requiring additional maneuvers. Furthermore, the policy exhibits strong generalization capabilities, successfully handling unseen objects. Importantly, it can be directly transferred to real robots without fine-tuning, consistently outperforming baseline methods. Videos of our experiments are available online: \href{https://sites.google.com/view/grasplargeflat}{\cblue{https://sites.google.com/view/grasplargeflat}}

\end{abstract}


\section{Introduction}
\label{sec:introduction}

\begin{figure}[!t]
      \centering
      \includegraphics[width=1.0\linewidth]{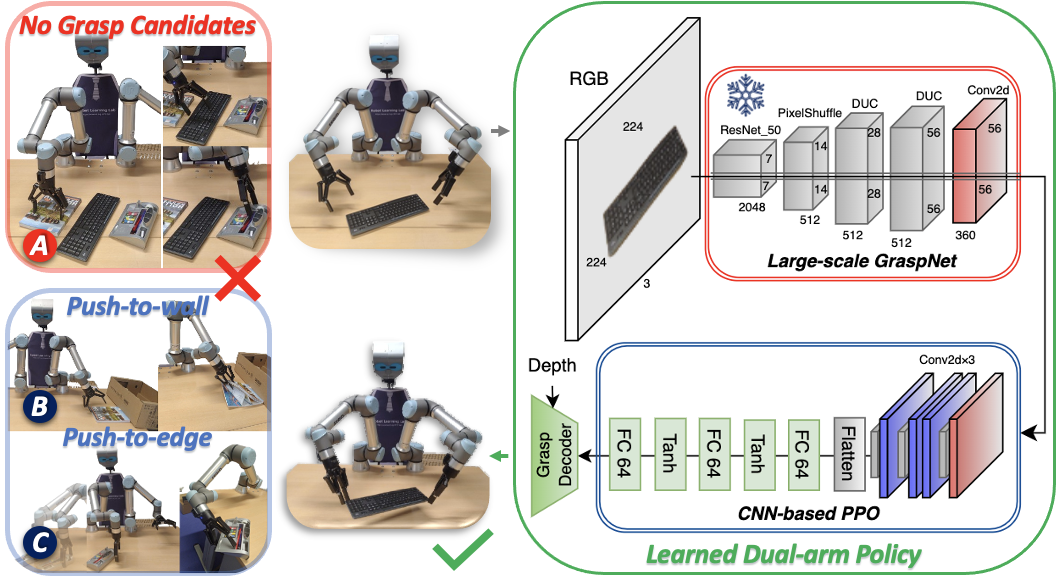}
      \vspace{-4mm}
      \caption{\textbf{Dual-Arm Coordinated Grasp Strategy for Large Flat Objects:} \textit{\textbf{A}} demonstrates that there are no efficient grasp candidates for large flat objects such as books and keyboards. \textit{\textbf{B}} and \textit{\textbf{C}} show current solutions: \textit{\textbf{B}} illustrates pushing the object against a wall to assist with grasping, while \textit{\textbf{C}} shows pushing the object to the table edge and grasping the overhanging part. However, \textit{\textbf{B}}'s method depends on using a wall, and \textit{\textbf{C}} is inefficient for dual-arm systems. To overcome these limitations, we propose a DRL framework to learn a cooperative dual-arm grasping strategy.}
      \label{fig:1}
\end{figure}

Consider a scenario where a large flat object, like a keyboard, lies flat on a table with a height that is lower than the gripper’s maximum stroke, making it ungraspable in its current state (see Fig.~\ref{fig:1} A) \cite{sarantopoulos2018human, do2024densetact, ma2023towards}. Human strategies often involve lifting the object into a semi-suspended position before grasping it from the side \cite{wu2023learning}. Pushing the object against a wall can also aid grasping but requires a nearby wall and a flat object edge (see Fig.~\ref{fig:1} B) \cite{liang2021learning}. Another method involves pushing the object to the table's edge to grasp it from the overhanging section (see Fig.~\ref{fig:1} C) \cite{zhang2023reinforcement}. While these approaches work well for single-arm robots, they are less suited for dual-arm systems, which can leverage the coordination of both arms for more sophisticated manipulation like humans without depending on environmental features \cite{hajj2023grasp}.

Recent research on single-arm robotic systems has introduced methods for grasping large flat objects by using pushing as a pre-grasp manipulation to create graspable poses \cite{li2022learning, ren2023learning, driess2020deep}. Three main approaches have emerged: visual push-grasping, push-to-wall, and push-to-edge. \textbf{Visual push-grasping} involves using pushing actions to separate objects in stacked scenarios. \cite{zeng2018learning} uses parallel DRL networks to jointly learn pushing and grasping policies, while \cite{xu2021efficient} introduces a goal-conditioned hierarchical RL framework for efficient push-grasping in cluttered environments. \cite{yu2023novel} combines transformers and CNNs, integrating a grasp detection network with a vision transformer for object positioning and a pushing transformer network. \textbf{Push-to-wall} involves pushing large objects against a wall to create a semi-hanging state suitable for grasping. \cite{liang2021learning} uses visual affordance maps and side-grasp action primitives for the Slide-to-Wall task, while \cite{sun2020learning} applies a DRL framework with Soft Actor-Critic (SAC) to teach a robot arm to push and lift flat, ungraspable objects. \textbf{Push-to-edge} involves pushing objects to the table's edge, creating an overhang that allows for easier grasping from the protruding part. \cite{zhang2023reinforcement} pre-trains a Variational Autoencoder to extract features from scenario images, and then uses PPO to learn pushing and grasping actions, moving objects to the table's edge before grasping. While these methods synergize pushing and grasping for task completion, they have notable limitations. Visual push-grasping relies on Deep Q-Networks (DQNs), which restricts flexibility. Push-to-wall uses a non-end-to-end approach with hard-coded grasping, weakening the interaction between pushing and grasping actions and struggling with beveled or smooth objects. Push-to-edge is time-consuming and effective only when objects can be easily pushed to the edge, limiting its applicability in more complex scenarios. Moreover, these methods fail to fully exploit the capabilities of dual-arm robotic systems, missing the potential advantages of coordinated use of both arms for more efficient and sophisticated manipulation \cite{yu2023coarse, kim2024goal, cui2024task}.

This paper proposes a model-free DRL framework aimed at optimizing the coordination between dual robotic arms for the purpose of grasping large flat objects. As illustrated in Fig.~\ref{fig:1}, we employ a large scale grasp pose detection model as the backbone to extract high-dimensional features from input RGB images, which are subsequently used as the state representation within a Markov Decision Process (MDP). We then design a specially designed Convolutional Neural Network (CNN)-based RL model, trained using PPO with shared Actor-Critic layers, to guide the policy in learning dual-arm cooperative grasp strategies. Experiments conducted in both simulated and real-world environments demonstrate that our system outperforms existing baselines and effectively accomplishes the task through smart dual-arm cooperation. Notably, our system seamlessly transitions from simulation to real-world applications without the need for additional data collection or fine-tuning.

The primary contributions of this paper are summarized as follows:

\begin{itemize}

    \item We propose a novel end-to-end RL framework designed to address the complex task of robotic manipulation, specifically for grasping large flat objects from challenging, ungraspable positions using vision-based inputs.  
    \item We validate the generalization ability of our policy across both familiar and novel objects, demonstrating its robustness. Furthermore, the policy successfully transfers from simulation to real-world robots without requiring any fine-tuning.
    \item To support and advance research in the field, we openly share our source code and pre-trained models, enabling reproducibility of our results and fostering further exploration by the research community.
    
\end{itemize}

\section{Related work}
\label{sec:related_work}

\subsection{Synergy of Pushing and Grasping}

The integration of pushing and grasping actions has been introduced to enhance object grasping in constrained scenarios, utilizing pushing actions to improve grasp quality \cite{yang2024attribute, newbury2023deep, li2023visual}. \cite{zeng2018learning} pioneered this approach with Visual Pushing for Grasping (VPG), a model-free DRL framework that learns joint policies of pushing and grasping through a parallel architecture, initiating the research on push-grasping synergy. \cite{xu2021efficient} proposed a goal-oriented hierarchical RL method with high sample efficiency, designed to learn push-grasping policies for specific object retrieval in cluttered environments. \cite{wang2024self} explored self-supervised learning for pushing and grasping in cluttered environments using a lightweight DRL model, demonstrating high resilience in managing complex scenes. \cite{yu2023novel} introduced a vision transformer-based pushing network (PTNet) and a cross-dense fusion network (CDFNet) for accurate grasp detection. Building on this, \cite{yu2024efficient} proposed an end-to-end push-grasping method using EfficientNet-B0 and a cross-fusion module. These studies focus on target-agnostic grasping methods, emphasizing general grasping strategies rather than object-specific solutions \cite{cao2024plot, gao2024improved}. 

The aforementioned methods are capable of managing certain challenging stacked scenarios, but their focus is limited to small objects. \cite{sun2020learning} and \cite{liang2021learning} extended grasping methods for large flat objects by first pushing them against a wall to create a semi-suspended state and then grasping from the side. \cite{sun2020learning} utilized DRL solely for predicting pushing actions and hard-coded the grasping step, while \cite{liang2021learning} employed DRL for pushing and used motion planning for grasping with a secondary robot. Both approaches are restricted to environments with walls. To address more complex objects like those with curved surfaces or irregular shapes, \cite{zhang2023reinforcement} proposed pushing large flat objects to the table edge before grasping the semi-hanging portion. Although this method avoids the need for walls, it is time-consuming and effective only when objects can be easily pushed to the edge, limiting its use in more complex scenarios.

\subsection{Reinforcement Learning for Robotic Manipulation}

Different robotic manipulation tasks employ various RL methods \cite{liu2024simulating}. For example, \cite{kasaei2023throwing} utilized Deep Deterministic Policy Gradient (DDPG) \cite{lillicrap2015continuous} and Soft Actor Critic (SAC) \cite{haarnoja2018soft} to train a robot for the task of throwing objects into a moving basket while avoiding obstacles. \cite{breyer2019comparing} developed a framework based on Trust Region Policy Optimization (TRPO) for object picking. For push-grasping, \cite{zeng2018learning}, \cite{xu2021efficient}, and \cite{wang2024self} used two parallel DQNs for pushing and grasping actions. Unlike these approaches, which rely on dual networks and are restricted to the DQNs method, our approach employs a single network for training. Moreover, the RL component in our method can be easily replaced with more efficient algorithms, offering greater flexibility.

\section{Method}
\label{sec:method}

\subsection{System Overview}
\label{subsec:syst}

\begin{figure*}[ht!]
    \centering
    \includegraphics[width=1\textwidth]{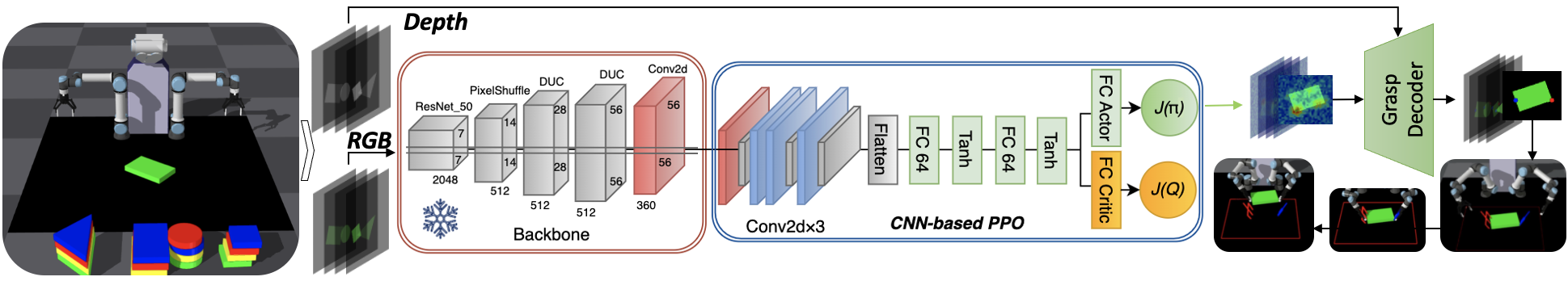}
    \caption{\textbf{System Overview of Dual-Arm Grasping Framework}: In the Isaac Gym environment, we integrate an RGB-D camera with a dual-arm UR5e robot. The camera captures RGB-D images and converts them into RGB-D top-down height maps. The RGB images are used as input observations for the framework, which employs a large-scale grasp network as the backbone to extract features. These features are fed into a CNN-based RL model trained using PPO. The output is a trained feature map, which is then decoded by a custom grasp decoder to generate executable actions within the environment. The decoder outputs two key points that determine the dual-arm grasp positions. During training, we utilize four distinct shaped objects.}
    \label{fig:2}
\end{figure*}

We model the dual-arm task of grasping large flat objects as an MDP within a hierarchical RL framework, enabling the agent to learn actions directly from visual inputs. The architecture, shown in Fig.~\ref{fig:2}, features an overhead camera mounted on the dual-arm robot that captures RGB-D images of the workspace. Each state $s_t$ consists of a color heightmap $c_t$ and a depth heightmap $d_t$, generated by projecting the original RGB-D images along the gravity direction, and a mask heightmap $g_t$ that identifies the target object within the workspace. The task is divided into two sub-tasks: first, a large-scale grasp pose detection model processes input RGB images; second, a custom CNN-based RL model, trained with PPO, outputs designed primitive motions. This approach uses visual cues for feature extraction and depth data for action validation, improving decision-making. Unlike existing methods, our framework focuses on learning cooperative strategies between the two arms, with human-like actions.

\subsection{Visual Representation}
\label{subsec:visu}

Large-scale models have gained widespread use across various fields. Inspired by GraspNet-1Billion \cite{fang2023robust, fang2023anygrasp}, the Angle-View Network (AVN), a large scale grasp pose detection model, decouples orientation into the approaching direction and in-plane rotation, framing it as a multi-class classification problem. Given its training on a large-scale dataset, our experiments demonstrate AVN's strong feature extraction capabilities, making it an ideal backbone for obtaining task-relevant compact representations from images. As illustrated in Fig.~\ref{fig:2}, the process starts with a ResNet50 encoder that converts the input image into high-dimensional feature representations. These features are subsequently decoded into next layers using a pixel-shuffle layer, followed by two Dense Upsampling Convolution (DUC) layers, which effectively reconstruct spatial orientation information.

\subsection{Learning Strategy}
\label{subsec:lear}

\subsubsection{State}
\label{subsubsec:state}

The state of the learning agent is determined by visual information from the environment, represented as a 3-channel RGB image. In our framework, this image acts as the observation for the MDP and has a resolution of $224 \times 224$ pixels. Consequently, the state $\bm{s}_t$ at time $t$ is defined as:

\begin{equation}
    \bm{s}_t = \text{RGB} \in \mathbb{R}^{3 \times 224 \times 224}
    \label{equ:3}
\end{equation}

\subsubsection{Action}
\label{subsubsec:action}

\begin{figure}[!t]
      \centering
      \includegraphics[width=1.0\linewidth]{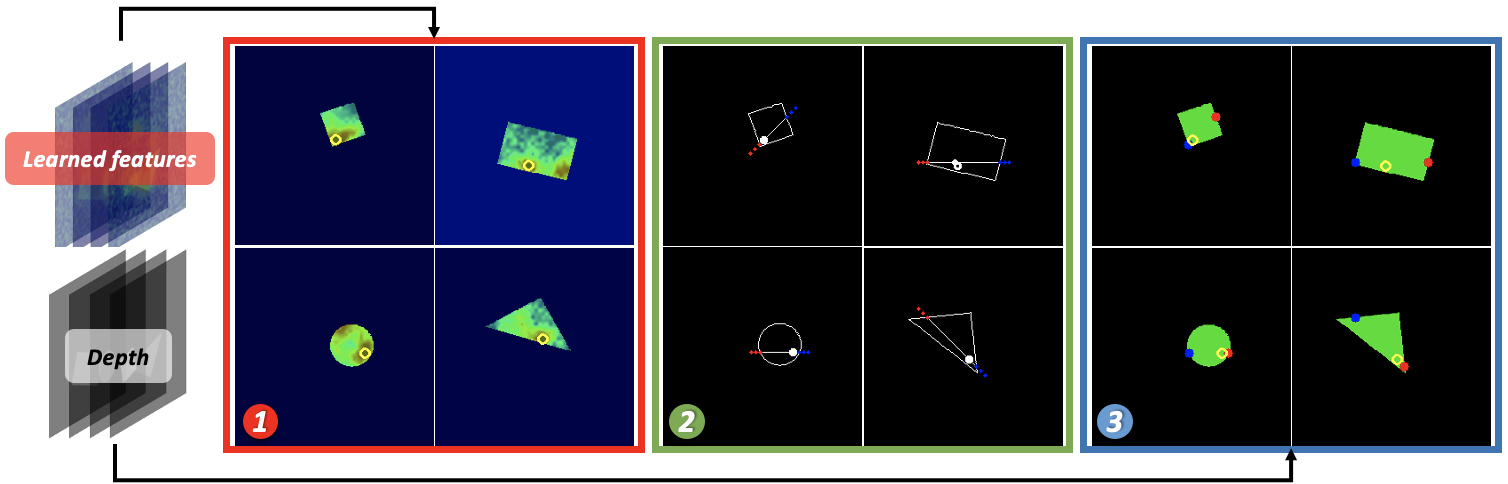}
      \caption{\textbf{Grasp Decoder for Dual-Arm Execution:} The Grasp Decoder converts the trained feature map from RGB images into actionable grasp points. First, the highest predicted pixel is selected (\textit{\textbf{a}}), then two points are aligned along one of 3 axes (\textit{\textbf{b}}) (If two points aren't found, alternatives are chosen hierarchically). The final action points (\textit{\textbf{c}}) guide the dual-arm robot’s grasp based on depth information.}
      \label{fig:3}
\end{figure}

The action output of the model is a feature map of size $56 \times 56$. We then design a decoder to select two symmetric points as the next grasping points, as illustrated in Fig.~\ref{fig:2}. The action representation is defined as follows:

\begin{equation}
    \bm{a}_t = \left[\bm{f}_{\text{map}} \right] \in \mathbb{R}^{56 \times 56}
    \label{equ:4}
\end{equation}
where, $\bm{f}_{\text{map}}$ denotes the feature map of size $56 \times 56$ obtained from the network's processing of the RGB input.

As illustrated in Fig.~\ref{fig:3}, the grasp decoder is designed to identify optimal grasping points on an object by utilizing visual features generated by our framework and depth data. In the initial step, the learned feature maps are analyzed, and the highest scoring point is selected as the main point of interest. In the second stage, the object's boundary is analyzed to identify the highest-scoring points along the contour. A line is then drawn through these points to detect intersections along the object's edge. The orientation of the line is determined through a hierarchical process, starting with the x-axis. If two operational points cannot be found or if the distance between the points is too short, the process moves to the next orientation. Finally, based on the line orientation and the identified intersections, the function calculates additional nearby points to refine the grasp positions, resulting in three specific points on each side of the object for precise grasping. These grasp points are then converted into real-world coordinates, accounting for the robot's configuration and workspace constraints.

\subsubsection{Reward}
\label{subsubsec:reward}

To learn robust and stable contact actions, after determining the grasp points for the dual-arm, the next step is for the dual-arm to lift the object to a height of $z = 0.45$ \textit{m} to evaluate the robustness of the action. Based on this, we design the reward function as follows:

\begin{equation}
    R = 
    \begin{cases}
        1, & \text{if grasp success}\\
        0, & \text{if not}
    \end{cases}
    \label{eq:greward}
\end{equation}

\subsubsection{Policy Optimization}
\label{subsubsec:policy}

We utilize a PPO policy with a CNN-based Actor-Critic architecture (see Fig.~\ref{fig:2}) to efficiently learn dual-arm coordination for grasping large flat objects. Given the use of two arms, non-prehensile actions are unnecessary; instead, our focus is on defining points that facilitate effective cooperation between the arms. The CNN-based RL model aids in learning an improved feature map, while the reward function guides the dual-arm agent toward learning robust actions.

\section{Experiments}
\label{sec:experiments}

This section details the experimental setup, and results of the manipulation tasks conducted in both simulation and real-world environments. Through these experiments, we aim to explore the following:

\begin{itemize}
    \item Are the primitive dual-arm motions generated by our policy more efficient, allowing tasks to be completed faster and more effectively than current state-of-the-art (SOTA) policies with single-arm robots?
    \item Can our policy be applied to seen and unseen large flat objects  even some irregular objects?
    \item Is the policy capable of transferring to a real robot without fine-tuning while outperforming existing baselines?
\end{itemize}

\subsection{Experimental Setups}
\label{subsec:experimental}

\begin{figure}[!t]
      \centering
      \includegraphics[width=1.0\linewidth]{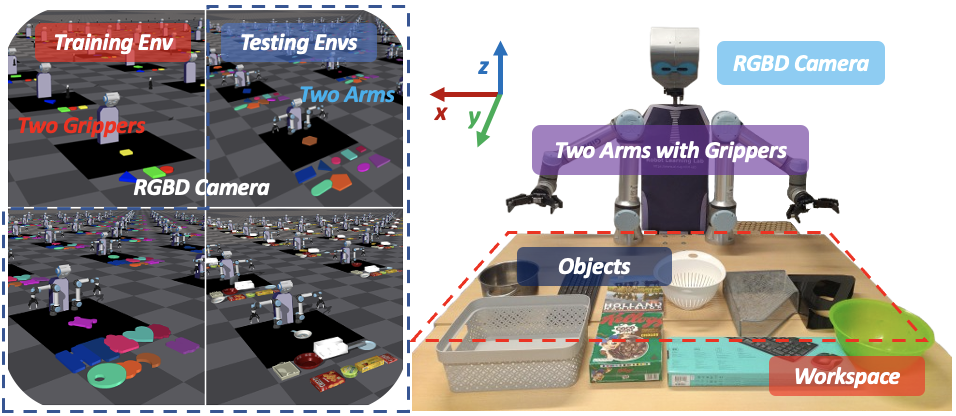}
      \caption{\textbf{Simulation and Real-World Scenarios:} In the simulation, objects are randomly placed in different positions and orientations, and the policy is trained in parallel environments within Isaac Gym. To accelerate training, a simplified version uses two grippers instead of full arms. To evaluate generalization, three object sets are tested: common large flat objects, irregularly shaped large flat objects, and household large flat objects. In the real-world scenario, the policy is validated on two UR5e robots equipped with Robotiq 2F-140 grippers.}
      \label{fig:4}
\end{figure}

We trained the proposed system using the Isaac Gym physics simulator and evaluated its performance in both simulated and real-world environments on a dual-arm robotic platform under various cluttered conditions. The simulated and real-world task setups are illustrated in Fig.~\ref{fig:4}. All experiments were performed on a desktop equipped with two Nvidia RTX 2080Ti GPUs, an Intel i7-9800X CPU, and 12GB of memory allocated per GPU.

\subsection{Simulation Experiments}
\label{subsec:simulation}

The simulated environment is constructed using Isaac Gym and includes two Universal Robots (UR5e) equipped with Robotiq 2F-140 grippers. RGB-D images are captured using a default camera setup. The robots operate in position-control mode, with push and grasp actions directly controlling the position of the end-effectors. Inverse kinematics is employed to convert these positions into joint space commands. The effective workspace measures $1.0 \times 1.0 \, m$, discretized into a grid of $224 \times 224$ cells, where each cell corresponds to one pixel in the orthographic image captured by the camera.

\subsubsection{Training Stage}
\label{subsubsec:training}

\begin{figure}[!t]
      \centering
      \includegraphics[width=1.0\linewidth]{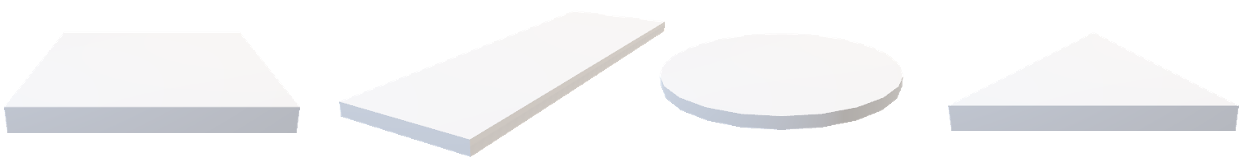}
      \par\noindent\makebox[0.23\linewidth][c]{\footnotesize Square} 
      \makebox[0.23\linewidth][c]{\footnotesize Rectangle} 
      \makebox[0.23\linewidth][c]{\footnotesize Circle} 
      \makebox[0.23\linewidth][c]{\footnotesize Triangle} 
      \caption{The training objects are divided into four categories of commonly shaped items, with each object randomly assigned a color during training.}
      \label{fig:5}
\end{figure}

The objects used for training are displayed in Fig.~\ref{fig:5}. We compare our method with several ablation variants to explore the following questions: 1) Is fine-tuning the pre-trained backbone necessary, and does it enhance performance? 2) Is it advantageous to use PPO with shared or dependent Actor-Critic layers?

\paragraph{Ablation Experiments}

\begin{figure}[!t]
      \centering
      \includegraphics[width=1.0\linewidth]{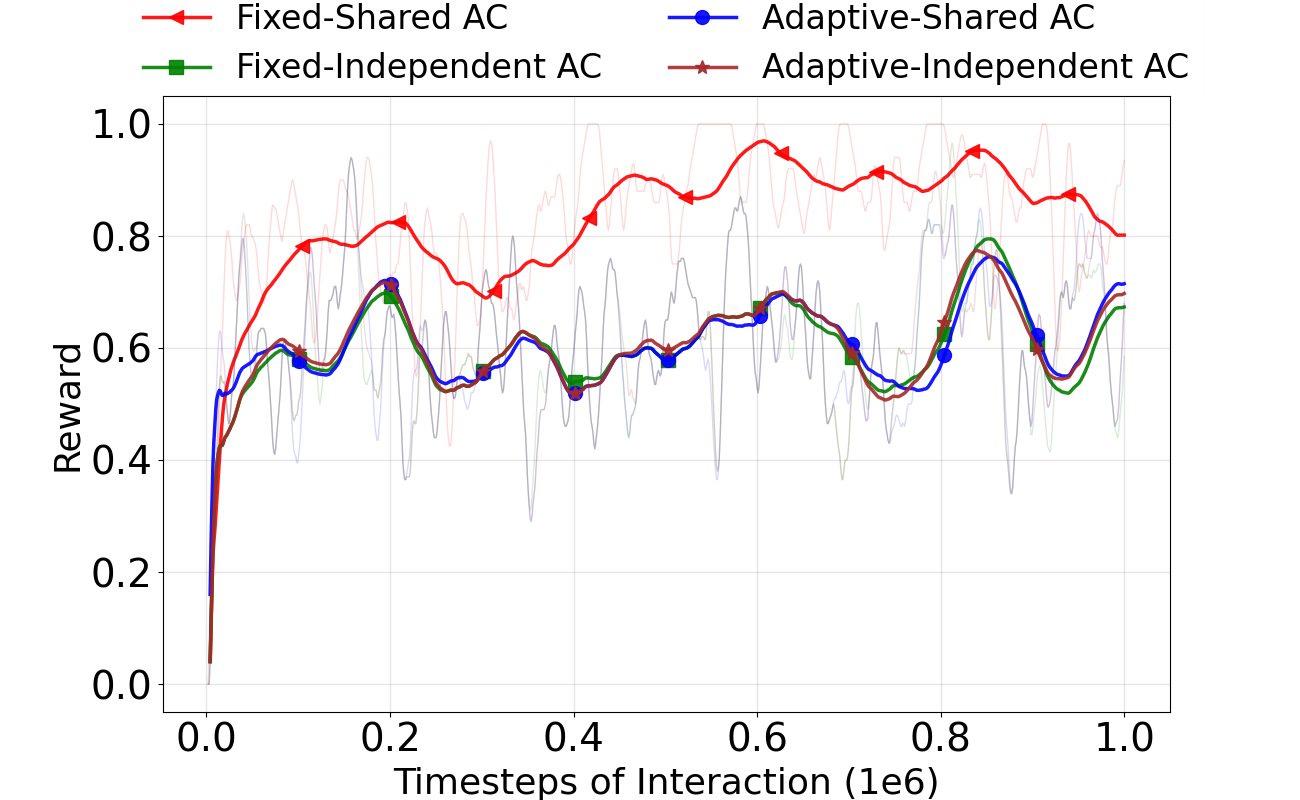}
      \caption{The ablation experiments explore how different configurations affect learning performance: examining the effects of keeping the pretrained backbone weights fixed (\textbf{Fixed Backbone}) versus allowing them to be updated during training (\textbf{Adaptive Backbone}), and comparing the impact of sharing neural network layers between the Actor and Critic (\textbf{Shared AC}) versus using separate layers (\textbf{Independent AC}).}
      \label{fig:6}
\end{figure}

In our approach, we use a large-scale deep learning grasp model as the backbone for image feature extraction. Leveraging this pre-trained model helps maintain a reasonable grasp success rate and establishes a strong feature map foundation at the outset, which is vital for RL models. However, this brings up an important question: should the pre-trained weights be kept frozen, or should they be fine-tuned to better adapt to our specific task? Additionally, in the CNN-based PPO, should the Actor-Critic layers be shared? How does this choice impact performance? Many existing methods utilize shared layers for their specific tasks to achieve improved results, making this a crucial aspect for experimental investigation. The results shown in Fig.~\ref{fig:6} reveal that freezing the backbone weights yields better performance than updating them continuously during training. Additionally, using PPO with shared Actor-Critic neural network layers achieves superior results.

\subsubsection{Testing Stage}
\label{subsubsec:testing}

\begin{figure}[!t]
      \centering
      \includegraphics[width=1.0\linewidth]{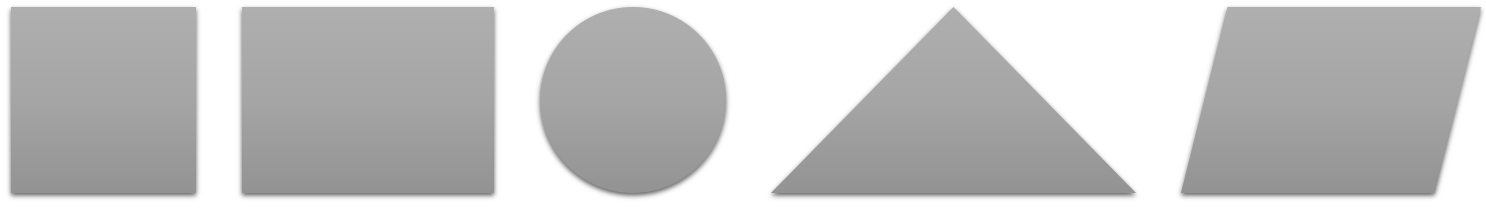}
      \par\noindent\makebox[0.18\linewidth][c]{\footnotesize Square} 
      \makebox[0.18\linewidth][c]{\footnotesize Rectangle} 
      \makebox[0.18\linewidth][c]{\footnotesize Circle} 
      \makebox[0.18\linewidth][c]{\footnotesize Triangle} 
      \makebox[0.18\linewidth][c]{\footnotesize Parallelogram} 

      \includegraphics[width=1.0\linewidth]{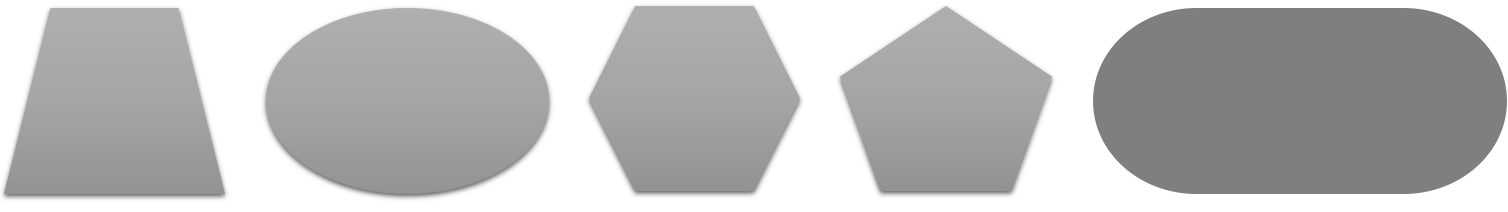}
      \par\noindent\makebox[0.18\linewidth][c]{\footnotesize Trapezoidal} 
      \makebox[0.18\linewidth][c]{\footnotesize Oval} 
      \makebox[0.18\linewidth][c]{\footnotesize Hexagonal} 
      \makebox[0.18\linewidth][c]{\footnotesize Pentagon} 
      \makebox[0.18\linewidth][c]{\footnotesize Notch} 

      \caption{\textbf{Large Flat Objects:} The objects used for testing experiments consist of ten categories of commonly shaped items.}
      \label{fig:7}
\end{figure}

\begin{figure}[!t]
      \centering
      \includegraphics[width=1.0\linewidth]{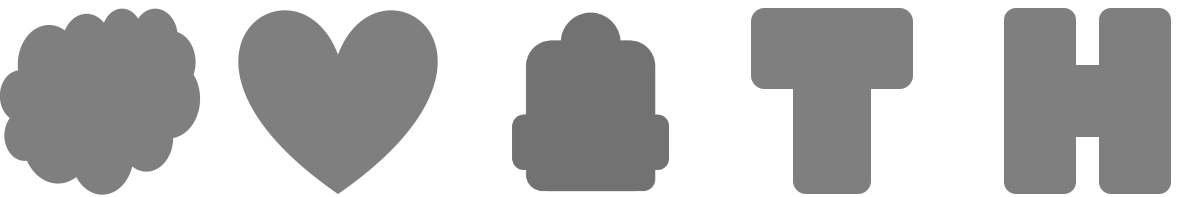}
      \par\noindent\makebox[0.18\linewidth][c]{\footnotesize (1)} 
      \makebox[0.18\linewidth][c]{\footnotesize (2)} 
      \makebox[0.18\linewidth][c]{\footnotesize (3)} 
      \makebox[0.18\linewidth][c]{\footnotesize (4)} 
      \makebox[0.18\linewidth][c]{\footnotesize (5)} 
      
      \includegraphics[width=1.0\linewidth]{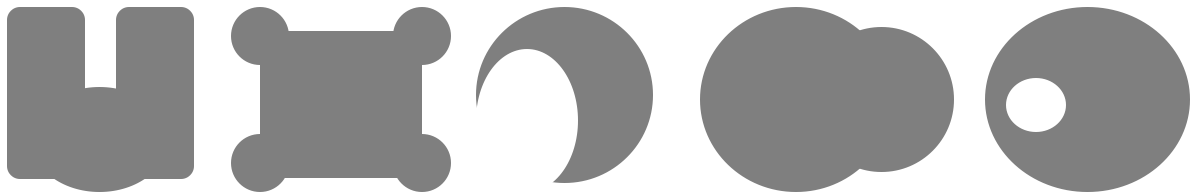}
      \par\noindent\makebox[0.18\linewidth][c]{\footnotesize (6)} 
      \makebox[0.18\linewidth][c]{\footnotesize (7)} 
      \makebox[0.18\linewidth][c]{\footnotesize (8)} 
      \makebox[0.21\linewidth][c]{\footnotesize (9)} 
      \makebox[0.18\linewidth][c]{\footnotesize (10)} 
      \caption{The irregularly shaped objects used for testing consist of letters and large-scale beveled objects, most of them with non-overlapping centers of mass and shape.}
      \label{fig:8}
\end{figure}

\begin{figure}[!t]
      \centering
      \includegraphics[width=1.0\linewidth]{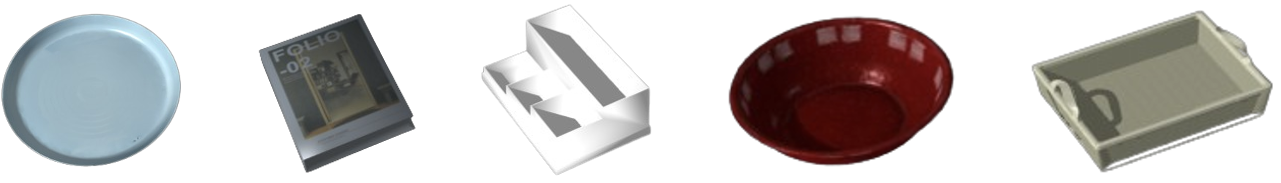}
      \par\noindent\makebox[0.18\linewidth][c]{\footnotesize Plate} 
      \makebox[0.18\linewidth][c]{\footnotesize Book} 
      \makebox[0.18\linewidth][c]{\footnotesize Bookholder} 
      \makebox[0.18\linewidth][c]{\footnotesize Bigbowl} 
      \makebox[0.18\linewidth][c]{\footnotesize Basket} 
      
      \includegraphics[width=1.0\linewidth]{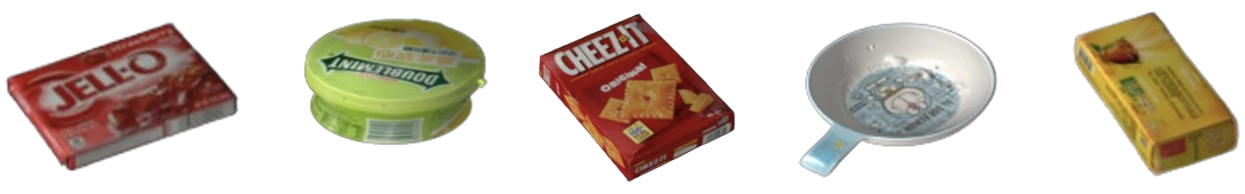}
      \par\noindent\makebox[0.18\linewidth][c]{\footnotesize Gelatinbox} 
      \makebox[0.18\linewidth][c]{\footnotesize Sugarcan} 
      \makebox[0.18\linewidth][c]{\footnotesize Crackerbox} 
      \makebox[0.18\linewidth][c]{\footnotesize Pot} 
      \makebox[0.18\linewidth][c]{\footnotesize Liptonbox} 
      \caption{The household objects used for testing include items from various scenarios, such as kitchen and office environments.}
      \label{fig:9}
\end{figure}

Generalization ability refers to how well a trained model can handle variations between training and testing data. To validate the generalization of our model, we trained our policy using three different types of large flat objects (see Fig.~\ref{fig:7} - Fig.~\ref{fig:9}). We evaluated the model’s generalization ability by testing it on objects with similar shapes but with bevels, as well as on new objects with irregular shapes and various household items. Each object was tested over 30 runs for comparison.

\paragraph{Comparison Experiments}
\noindent

\begin{table*}[!t]
    \begin{center}
        \caption{Results for 10 large flat objects with bevels (Mean \%)}
        \resizebox{\linewidth}{!}{
        \begin{tabular}{ c c c c c c c c c c c c }
        \hline
        Shape & Square & Rectangle & Circle & Triangles & Parallelogram & Trapezoidal & Oval & Hexagonal & Pentagon & Notch & All \\
        \hline
        
        Push-to-edge & 90.0 & 86.7 & 96.7 & 80.0 & 86.7 & 86.7 & 96.7 & 90.0 & 93.3 & 90.0 & 89.7 \\
        
        Ours & \textbf{96.7} & \textbf{96.7} & 96.7 & \textbf{93.3} & \textbf{90.0} & \textbf{90.0} & 96.7 & \textbf{93.3} & \textbf{96.7} & \textbf{93.3} & \textbf{94.3} \\
        
        \hline 
        
    \end{tabular}}
    \label{table:1}
    \end{center}
\end{table*}

\textbf{Large Flat Objects:} Grasping large flat objects with bevels presents a significant challenge, as traditional methods struggle due to the lack of sufficient space for the gripper to push the object against a wall to achieve a hanging state. Additionally, the push-to-edge strategy is not compatible with our dual-arm system. In contrast, our approach leverages two arms to identify two operable points, allowing the object to be lifted using these learned grasping points. We evaluated our best-performing model from the ablation studies on ten categories of objects to assess its generalization performance, as shown in Table.~\ref{table:1}. The results demonstrate that our method outperforms others, consistently achieving a success rate of over $90\%$ across all beveled objects. These findings indicate that our model effectively generalizes to beveled objects.

\begin{table*}[!t]
    \begin{center}
        \caption{Results for 10 irregularly shaped large flat objects (Mean \%)}
        \resizebox{\linewidth}{!}{
        \begin{tabular}{ c c c c c c c c c c c c }
        \hline
        Object ID & Object 1 & Object 2 & Object 3 & Object 4 & Object 5 & Object 6 & Object 7 & Object 8 & Object 9 & Object 10 & All \\
        \hline
        
        Push-to-edge & 100.0 & 83.3 & 96.7 & 83.3 & 90.0 & \textbf{96.7} & 80.0 & 83.3 & 80.0 & 100.0 & 89.3 \\
        
        Ours & \textbf{100.0} & \textbf{90.0} & 96.7 & \textbf{93.3} & \textbf{93.3} & 93.3 & \textbf{90.0} & \textbf{96.7} & \textbf{100.0} & \textbf{100.0} & \textbf{95.3} \\
        
        \hline 
        
    \end{tabular} }
    \label{table:2}
    \end{center}
\end{table*}

\textbf{Irregularly Shaped Large Flat Objects:} To further evaluate the generalization capability of our model, we introduced new and more challenging objects with irregular shapes, as depicted in Fig.~\ref{fig:8}. The results of these tests are presented in Table.~\ref{table:2}. Our model consistently maintained a high success rate, demonstrating robust performance across different object shapes. In comparison, the push-to-edge method requires a portion of the object to be pushed off the table, which limits its effectiveness, especially for irregularly shaped objects. While the performance of some irregular objects surpassed that of beveled objects, the bevels can alter the gripper’s movement, leading to discrepancies between the intended and actual grasp positions. Despite these challenges, our model exhibited similar performance across all object types, highlighting its robustness and adaptability. This demonstrates that our policy effectively handles new objects, even those with highly irregular shapes.

\begin{table*}[!t]
    \begin{center}
        \caption{Results for 10 Household Large Flat Objects (Mean \%)}
        \resizebox{\linewidth}{!}{
        \begin{tabular}{ c c c c c c c c c c c c }
        \hline
        Object ID & Plate & Book & Bookholder & Bigbowl & Basket & Gelatinbox & Sugarcan & Crackerbox & Pot & Liptonbox & All \\
        \hline
        
        Push-to-edge & 73.3 & 76.7 & 70.0 & \textbf{86.7} & 83.3 & 86.7 & 73.3 & 86.7 & 66.7 & 80.0 & 78.3 \\
        
        Ours & \textbf{96.7} & \textbf{80.0} & \textbf{86.7} & 83.3 & 83.3 & 86.7 & 73.3 & 86.7 & \textbf{83.3} & \textbf{86.7} & \textbf{84.7} \\
        
        \hline 
        
    \end{tabular} }
    \label{table:3}
    \end{center}
\end{table*}

\textbf{Household Large Flat Objects:} To assess the generalization capability of our model in real-world scenarios, we evaluated its performance on 10 large household flat objects, as shown in Fig.~\ref{fig:9}. The outcomes of these tests are summarized in Table~\ref{table:3}. The results demonstrate that our model remains robust and effective, even when faced with previously unseen objects. This is largely due to the use of a pre-trained backbone, which allows the model to maintain high efficiency with untrained items.

\subsection{Real-World Experiments}
\label{subsec:real}

\begin{figure}[!t]
      \centering
      \includegraphics[width=1.0\linewidth]{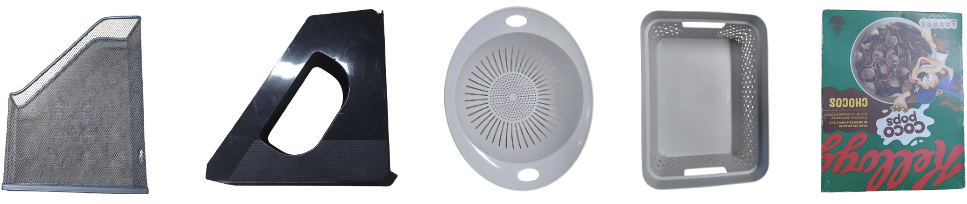}
      \par\noindent\makebox[0.18\linewidth][c]{\footnotesize Bookholder1} 
      \makebox[0.18\linewidth][c]{\footnotesize Bookholder2} 
      \makebox[0.18\linewidth][c]{\footnotesize Washbasket} 
      \makebox[0.18\linewidth][c]{\footnotesize Storagebasket} 
      \makebox[0.18\linewidth][c]{\footnotesize Cerealbox} 
      
      \includegraphics[width=1.0\linewidth]{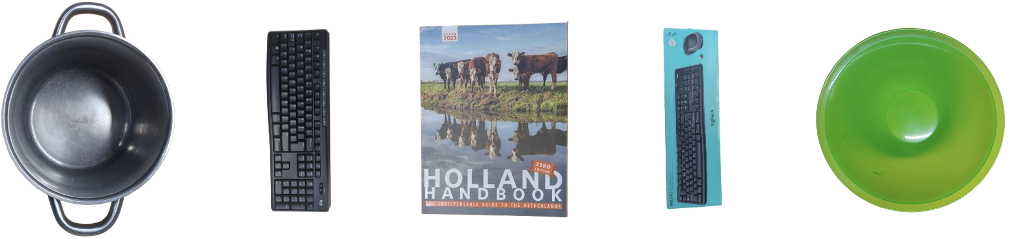}
      \par\noindent\makebox[0.18\linewidth][c]{\footnotesize Steelpot} 
      \makebox[0.18\linewidth][c]{\footnotesize Keyboard} 
      \makebox[0.18\linewidth][c]{\footnotesize Book1} 
      \makebox[0.18\linewidth][c]{\footnotesize Box} 
      \makebox[0.18\linewidth][c]{\footnotesize Greenbowl} 
      \caption{The real robot experiments comprises various types of everyday objects, with widths ranging from 180 mm to 440 mm and heights from 25 mm to 95 mm.}
      \label{fig:10}
\end{figure}

\begin{table*}[!t]
    \begin{center}
        \caption{Results for 10 objects in real-world (Mean \%)}
        \resizebox{\linewidth}{!}{
        \begin{tabular}{ c c c c c c c c c c c c }
        \hline
        Object ID & Bookholder1 & Bookholder2 & Washbasket & Storagebasket & Cerealbox & Steelpot & Keyboard & Book1 & Box & Greenbowl & All \\
        \hline

        Push-to-edge & 80.0 & 86.7 & 66.7 & 73.3 & 86.7 & 73.3 & \textbf{86.7} & 73.3 & 86.7 & 80.0 & 79.3 \\
        
        Ours & \textbf{93.3} & 86.7 & \textbf{100.0} & \textbf{93.3} & \textbf{100.0} & \textbf{80.0} & 66.7 & 73.3 & \textbf{93.3} & \textbf{86.7} & \textbf{87.3} \\
        
        \hline 
        
    \end{tabular} }
    \label{table:4}
    \end{center}
\end{table*}

We validated our system through real-world experiments using a physical setup comprising two UR5e robotic arms equipped with Robotiq 2F-140 grippers, and an Asus Xtion camera capturing RGB-D images. As shown in Fig.~\ref{fig:10}, our model was tested on various large flat objects. In these experiments, each test scenario was repeated $n = 15$ times. Notably, our models were transferred directly from simulation to the real-world environment without any additional retraining. The detailed results, including the grasp success rate for each object, are presented in Table~\ref{table:4}. These outcomes highlight our policy’s strong generalization capabilities to real-world settings. Furthermore, we tested the model on novel objects that were not included in the training phase, demonstrating that our approach can successfully adapt to previously unseen items. For further details, please visit \href{https://sites.google.com/view/grasplargeflat}{\cblue{https://sites.google.com/view/grasplargeflat}}.

\subsection{Other Challenging Scenes}

\begin{figure}[!t]
      \centering
      \includegraphics[width=1.0\linewidth]{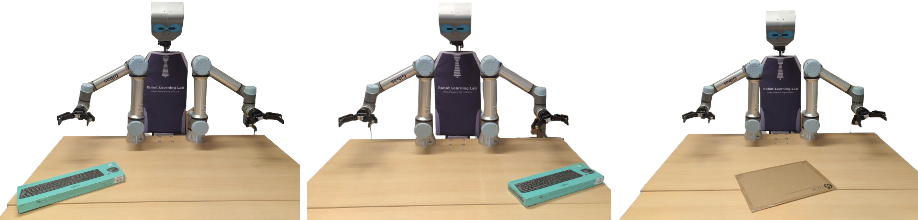}
      \vspace{-4mm}
      \caption{Three challenging scenes are presented where traditional grasping methods fail. From left to right: in the left and middle scenes, the objects are positioned off-center, either biased to the left or right, requiring a slight push towards the center for successful grasping. In the right scene, the object is too thin, making it difficult to grasp effectively.}
      \label{fig:11}
       \vspace{-4mm}
\end{figure}

Despite the success of our dual-arm grasping strategy, some limitations remain. As illustrated in Fig.~\ref{fig:11}, 
one particularly difficult scenario involves objects that are positioned off-center (\textit{left} and \textit{middle} images). A potential solution would be to integrate a pre-grasp maneuver, such as sliding or pushing the object slightly to center it before executing the dual-arm grasp. Another challenging case involves extremely thin or flat objects, which can be difficult for the robot to pick up due to the lack of sufficient grip surface. 

\subsection{Failure Cases}

We identified three failure types during the experiments:

\textbf{Asynchronous Dual-Arm Motion Challenges:} A key challenge is the asynchronous movement of the arms, disrupting coordination and reducing grasping precision. Future work should focus on synchronizing arm motions.

\textbf{Handling Objects with Mixed Rigidity (e.g., Half Hard, Half Soft):} Grasping objects with mixed properties, like books with flexible covers, challenges the dual-arm policy to adapt to varying materials. While our framework shows adaptability, further refinement is needed to better handle complex textures and varying rigidity.

\textbf{Limitations of Position Control and Lack of Force Feedback:} Position control without force sensors can cause unintended damage to rigid objects due to the lack of feedback on excessive forces. This limitation underscores the need for integrating force or tactile feedback, enabling safer handling of fragile or rigid items. Future work should focus on incorporating force-sensing to improve dual-arm coordination and robustness in real-world applications.

\section{Conclusion}
\label{sec:conclusion}

In this paper, we presented a novel framework for dual-arm coordination aimed at grasping large flat objects that are otherwise ungraspable by single-arm systems. By leveraging a large-scale grasp pose detection model as the backbone for feature extraction and integrating it with a CNN-based PPO algorithm, we successfully enabled coordinated, human-like grasping strategies. Our method demonstrates significant improvements over SOTA approaches in simulation and real-world environments, showcasing the effectiveness of dual-arm cooperation for complex manipulation tasks. The extensive experiments conducted on a diverse range of objects, from standard flat surfaces to irregular and previously unseen household items, demonstrate the robustness and adaptability of our approach. The results show that our method achieves a higher success rate under the same testing conditions compared to other approaches. However, our method has some limitations, such as difficulty in handling off-center or extremely thin objects. Future work will focus on enhancing pre-grasp manipulation techniques and incorporating dynamic repositioning strategies, to further improve the system's versatility in challenging environments.




\bibliographystyle{ieeetr}
\bibliography{reference}




\end{document}